# Validated Data Onboarding for AI Demand Forecasting on U.S. Building Meter Data: Design, Controlled Evaluation, and a Corrected Negative Result

Yixuan Liang[1] · [1]Independent researcher; M.S. Computer Science, Illinois Institute of Technology, Chicago (2022). Software engineer working on large-scale data onboarding and AI infrastructure (Meta Platforms; ByteDance/TikTok).
Contact: liangyixuan333@gmail.com
Technical report. Code, data manifests and all result bundles: github.com/LIANGYIXUAN3335/energy-data-onboarding-forecasting (MIT license).

**Abstract.** Electric utilities and grid operators increasingly rely on machine-learning models to forecast next-day demand, and those models learn from meter data that is routinely defective: readings go missing, sensors freeze, buildings read zero for hours, and units change by a factor of 100. This report presents a data-onboarding pipeline that detects and repairs such defects before a model is trained, using only information available at forecast time, and a controlled experiment that measures whether the pipeline protects a 24-hour-ahead forecast. On hourly electricity data for twelve U.S. buildings from the public Building Data Genome 2 dataset (210,528 rows, 2016–2017), seeded, hash-logged defects touching 0.10% of the training period raised the error of a gradient-boosting forecaster by 86%; after detection and past-only repair the error returned to the clean-data level (mean absolute scaled error 0.760 clean, 1.415 corrupted, 0.729 repaired) while 93% of training targets were retained. At a defect prevalence calibrated to published field studies (1.6% of training rows) the unprotected forecaster's error reached 4.4 times that of a seasonal-naive rule, and the repaired forecaster again matched the clean baseline. The same pattern held for ridge regression and a random forest and across horizons of 1 to 24 hours. A first version of the pipeline over-cleaned natural data and made forecasts 25% worse; that result is retained, its cause is traced in the published artifacts, and the per-building calibration that corrects it is documented as a dated amendment. Every number is reproducible from pinned public inputs with SHA-256 verification, 84 automated tests and continuous integration.

## 1. Motivation

The U.S. electric grid is being asked to absorb rapid load growth from data centers, electrification and manufacturing while integrating variable renewable generation. The Department of Energy's Genesis Mission lists "Scaling the Grid to Power the American Economy" among its national science and technology challenges, citing reliability limits under demand growth from data centers, manufacturing and electrification [1]; the 2025 America's AI Action Plan calls for world-class, AI-ready scientific datasets with minimum data-quality standards, treating such data as a national strategic asset [2]. Demand forecasting is one of the first places AI is applied in grid operations, and its inputs are meter streams whose quality is far from guaranteed. Field studies report that up to 4% of readings are missing in typical smart-meter deployments and far more in some panels, that zero readings affect 1–6% of intervals or customer-days, and that unit and multiplier errors affect a minority of meters and records (Section 2). A forecaster trained on such data silently learns the defects.

The question this work answers is narrow and testable: *can an onboarding pipeline detect and repair meter-data defects before a model is trained, using only information that would have been available at the time, and does that measurably protect the forecast?* The contribution is not a new forecasting model. It is (a) a pipeline whose checks are calibrated per meter on clean history, whose repairs never read a later value, and whose every artifact is bound to its inputs by cryptographic hashes; (b) a three-condition evaluation design (clean, corrupted, repaired) that isolates the effect of the pipeline from the effect of the model; and (c) an honest record of a first version that failed, why it failed, and how the failure was diagnosed from the pipeline's own outputs.

## 2. How common are meter-data defects?

To choose a realistic corruption level, published prevalence figures were compiled (the full table with sources is in `docs/fault_prevalence.md` of the repository). Missing readings dominate: 0.03% in the Irish CER trial and 3.5% in Korean residential data [4], 0.17% on average in the Low Carbon London pricing trial [11], and 3–4% in deployed smart-meter systems as summarized by Weber et al. [3]. In the British SERL panel, 17.2% of expected half-hourly electricity reads were not received [5]. In three example U.S. Army buildings analyzed by Pacific Northwest National Laboratory, missing timestamps amounted to about 2.6%, 7.5% and 30.5% of expected readings, and in the two worse buildings most of the missing data fell in gaps longer than a week [6]. At utility scale, more than 93% of 178,333 metered premises missed fewer than 2% of intervals; the counts in that study's Table 1 imply that 2.1% of premises read zero for an entire year and 0.08% carried extreme spikes [7]. Zero values made up 6.2% of readings in another example building [6] and appeared in 1.04% of customer-days in a residential program [8]. Unit errors are also common: in the SERL panel 7.2% of expected daily electricity reads (about 9% of those received) were flagged as likely recorded in the wrong unit [5], and 1.05% of records in a residential dataset were duplicates differing by a multiple consistent with a meter multiplier [8]. Manual annotation of 1,413 electricity meters from the ASHRAE Great Energy Predictor III competition, whose data come from the same building collection as the dataset used here, labels 1.66% of non-missing readings as anomalous [9, 10]. These figures motivate two experimental densities: a sparse suite of four events per fault family (0.10% of training rows) to isolate mechanisms, and a dense suite of 1.6% of training rows at the lower end of the field reports.

## 3. System design

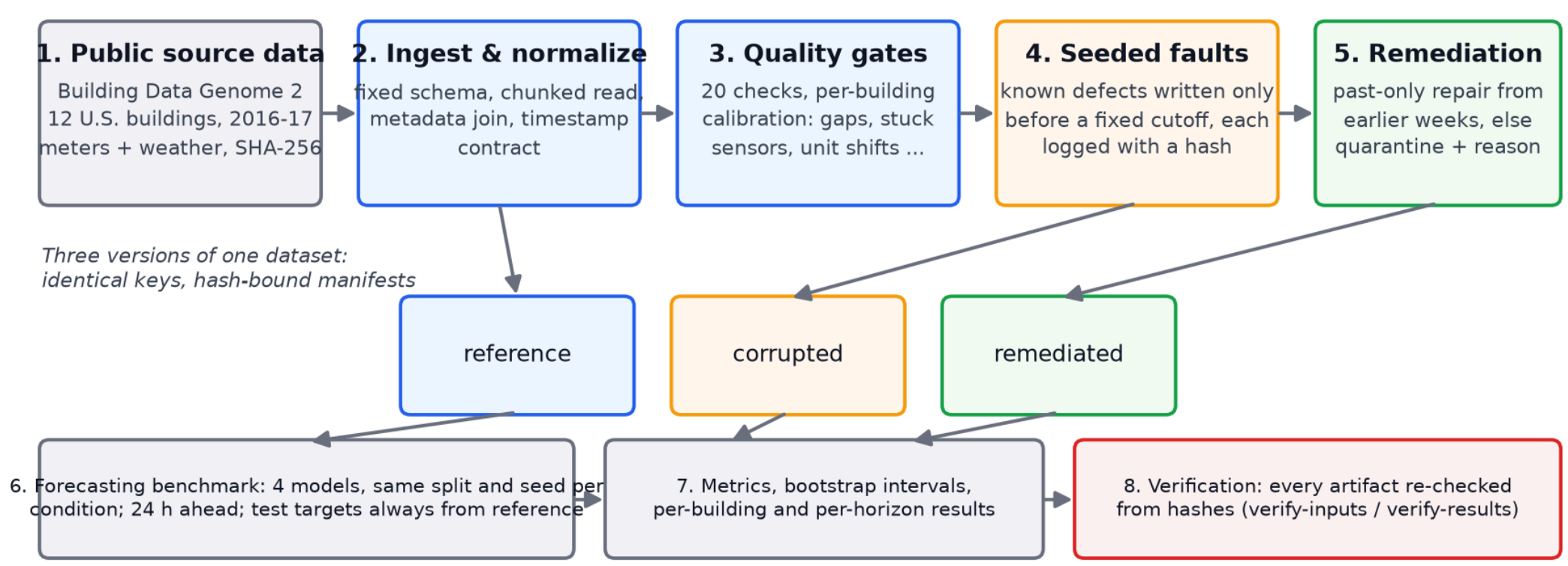

**Figure 1.** The pipeline. Public source files are verified against pinned hashes, normalized, checked by twenty quality gates calibrated per building, corrupted by seeded and logged faults inside the training period only, repaired with past-only rules, and published as three hash-bound conditions that feed one forecasting benchmark and one verifier.

**Source pinning.** The three input files of Building Data Genome 2 v1.0 [10] (electricity meters, building metadata, site weather) are pinned by URL, byte size and SHA-256 in a provenance manifest; a download that does not match is refused and no run can start without the verification record.

**Contract and checks.** Every condition file has exactly six columns (`timestamp, building_id, load, site_id, primary_use, condition`) and one row per building-hour. Twenty named checks cover schema and timestamp validity, duplicate keys, missing, infinite and negative loads, robust outliers against a causal 168-hour median, long zero runs, stuck (constant) readings, level or unit shifts, metadata integrity and coverage of the train/test boundary. Structural failures always block publication; value-quality checks carry explicit count or rate thresholds recorded in the configuration.

**Per-building calibration.** The stuck-sensor rule raises its 8-hour threshold, per building, to just above the 0.999 quantile of that building's natural constant-run lengths, computed on the clean condition strictly before the experiment cutoff and then frozen for every condition; a zero run counts only during hours when the building is normally active (its causal hour-of-week median over the previous eight weeks exceeds a quarter of its trailing weekly median), and the decision is taken for the run as a whole. The applied thresholds are published in `detector_calibration.csv`.

**Seeded faults.** Six fault families are written into the training period only: single missing, negative and spike values, and 8-hour blocks of zeros, of a constant value and of a ×100 unit error, the blocks placed only over strictly positive readings so that every seeded fault is observable. Each affected key is logged with the original and corrupted value and a per-event digest. Rows at or after the cutoff are never touched, and a consumer-side check refuses any bundle in which they differ from the clean condition.

**Past-only repair.** Gaps of up to three hours are forward-filled from the last valid reading; gaps of up to 48 hours take the median of the same weekday and hour over the previous eight weeks, computed from earlier, originally valid readings only; a flagged zero is never filled with a zero; suspected unit shifts receive the profile value rather than being divided by an inferred factor; anything else is quarantined as a null with a reason code. No repair reads a later value.

**Provenance and verification.** Each run writes manifests with the hashes of its inputs, configuration, source tree and outputs. The forecasting package's `verify-inputs` command re-checks the producer bundle (hashes, equal keys, every changed value against the fault log, the cutoff) before any model is fitted; `verify-results` recomputes the metrics, intervals and figures from the stored predictions, binds the targets back to the producer bundle, and checks the declared model set against the one the caller expects.

## 4. Experimental protocol

**Data.** Twelve buildings (two education and two office buildings at each of three U.S. sites in the Eastern time zone, identified as Eagle, Panther and Rat in the source dataset), hourly electricity from January 2016 to December 2017, 210,528 rows. The selection was declared before modeling and does not inspect forecast accuracy.

**Split and conditions.** Training through December 31, 2016; a validation partition (2017 Q1) reserved and unused because models and hyperparameters were fixed a priori; test April to December 2017 (79,178 hourly targets). Faults are seeded only before the cutoff of January 1, 2017. Three conditions of the training data (reference, corrupted, remediated) are evaluated on identical test hours whose true values always come from the reference data, so the conditions differ only in what the model learned.

**Task, features and models.** For a target hour $t$, every load- and weather-derived feature is observed at $t-24$ h or earlier: load lags of 24, 48, 168 and 336 hours; trailing 24- and 168-hour statistics ending at $t-24$; hour, weekday, day-of-year and month cycles; weekend and U.S. federal holiday flags; and site air temperature, dew point and wind speed at $t-24$ with lagged trailing means (30 features). No weather forecast is used. Four models with fixed hyperparameters are trained on each condition: a seasonal-naive rule (same hour last week, then yesterday), ridge regression, a random forest (200 trees) and histogram gradient boosting (120 iterations). Preprocessing is fitted on training rows only.

**Metrics.** The primary metric is pooled mean absolute scaled error (MASE) with a 168-hour seasonal scale computed once from clean training data, so 1.0 equals the seasonal-naive rule; supporting metrics are MAE with row-level bootstrap intervals (500 repetitions), RMSE, weighted MAPE, $R^2$, per-building MASE, and paired remediated-minus-corrupted absolute-error differences on identical keys. A separate direct multi-horizon run reports horizons 1–24 with a building-week block bootstrap.

**Pre-specification and amendments.** Split dates, features, models, hyperparameters and metrics were written down in a protocol file before the runs they govern. The second version of the pipeline (v3) was designed after the first result was examined and is documented as a dated amendment; both versions' result bundles are kept and verified.

## 5. Results

### 5.1 The corrected experiment (v3)

**Table 1.** Pooled MASE on 79,178 test hours (April–December 2017), targets always from clean data. Sparse fault suite: 24 events, 108 of 105,408 training rows; 95% intervals from 500 row-level bootstrap repetitions. Paired differences are remediated minus corrupted absolute error in kWh; negative favors repair.

| Model | Reference (clean) | Corrupted | Remediated | Paired MAE difference (95% CI) |
|---|---|---|---|---|
| Gradient boosting | 0.760 | 1.415 (+86%) | **0.729** | −42.5 (−44.4 to −40.6) |
| Random forest | 0.741 | 1.040 (+40%) | **0.715** | −20.1 (−21.2 to −19.1) |
| Ridge regression | 0.812 | 1.192 (+47%) | **0.809** | −23.7 (−24.2 to −23.3) |
| Seasonal naive | 1.040 | 1.040 | 1.040 | 0 (uses no training data) |

Undetected defects hurt: 108 corrupted rows raised gradient-boosting error by 86%, because one model is shared by all twelve buildings and a single ×100 segment or zero block distorts what it learns for every building (the office building that received two ×100 segments went from 36 to 251 kWh MAE, and buildings with no block faults also worsened). After detection and past-only repair every learned model returned to its clean-data error. The repaired models score slightly below the clean baseline; that further gain is observational, not controlled: it arises because the pipeline also quarantined or repaired about 9,100 rows of natural 2016 anomalies, most of them long zero stretches at one site, and it retains 93.0% of the training targets.

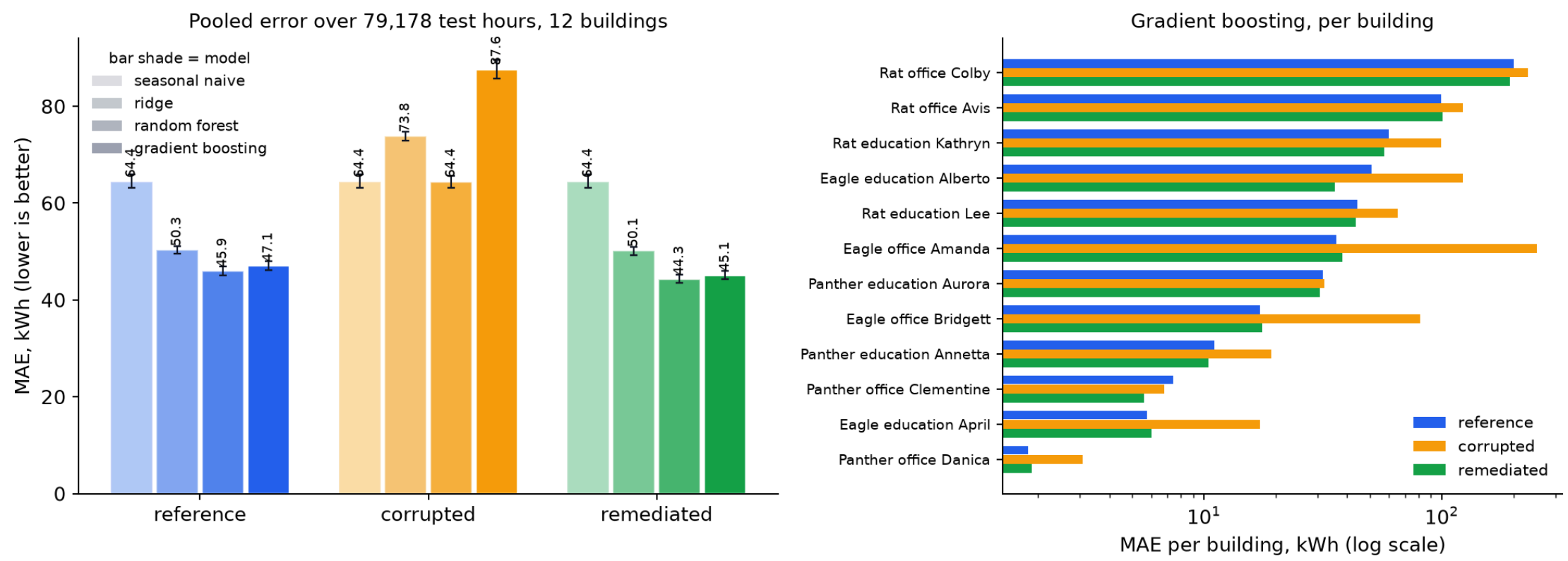

**Figure 2.** Left: pooled MAE by condition and model with bootstrap intervals. Right: gradient-boosting MAE per building on a log scale; the buildings that received block faults show the largest corrupted-condition error, and the repaired bars sit at or below the clean bars.

### 5.2 At the defect prevalence the literature reports

**Table 2.** Same protocol with 855 seeded events on 1,660 training rows (1.6%), calibrated to the field studies of Section 2.

| Model | Reference (clean) | Corrupted (1.6% of rows) | Remediated | Paired MAE difference (95% CI) |
|---|---|---|---|---|
| Gradient boosting | 0.760 | 4.545 | **0.729** | −236 (−246 to −225) |
| Random forest | 0.741 | 3.393 | **0.711** | −166 (−173 to −160) |
| Ridge regression | 0.812 | 2.843 | **0.806** | −126 (−127 to −125) |

At this density an unprotected forecaster is 2.7 to 4.4 times worse than the seasonal-naive rule, and the repaired models land exactly where they did at the sparse density while keeping 95.8% of the training targets. Detector recall at this density is 1.00 for missing, negative and unit-scale faults, 0.98 for zero blocks, 0.93 for spikes and 0.68 for stuck segments; precision for missing values rises to 0.92 because natural false positives are diluted by true detections.

### 5.3 Horizons 1 to 24 hours

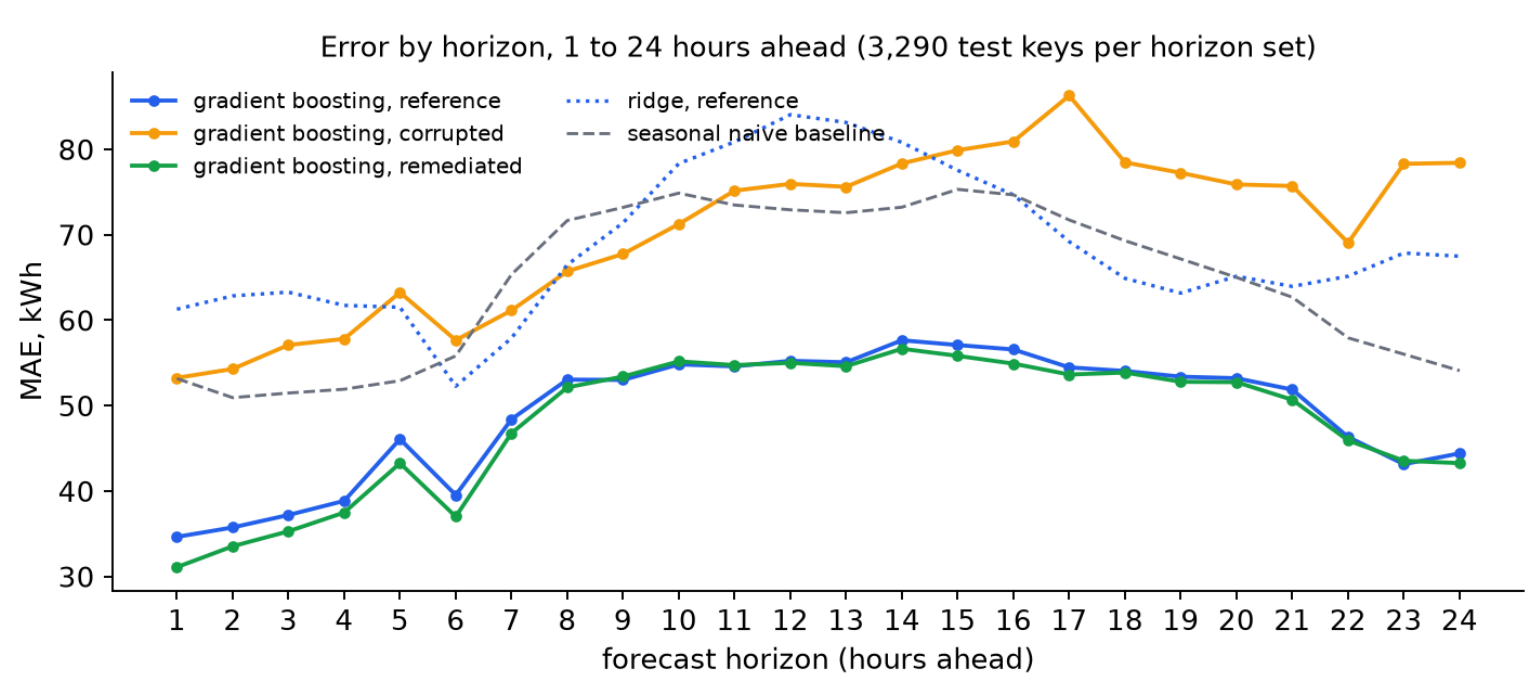

**Figure 3.** Direct multi-horizon sensitivity (one origin per day at 23:00, 78,960 test keys, a 22-feature direct-forecast set with weather observed at the origin): gradient-boosting MAE by horizon for the three conditions, with ridge and the seasonal-naive rule on clean data. Pooled MASE 0.793 clean, 1.140 corrupted, 0.776 repaired. The ordering holds at every horizon.

### 5.4 One building, step by step

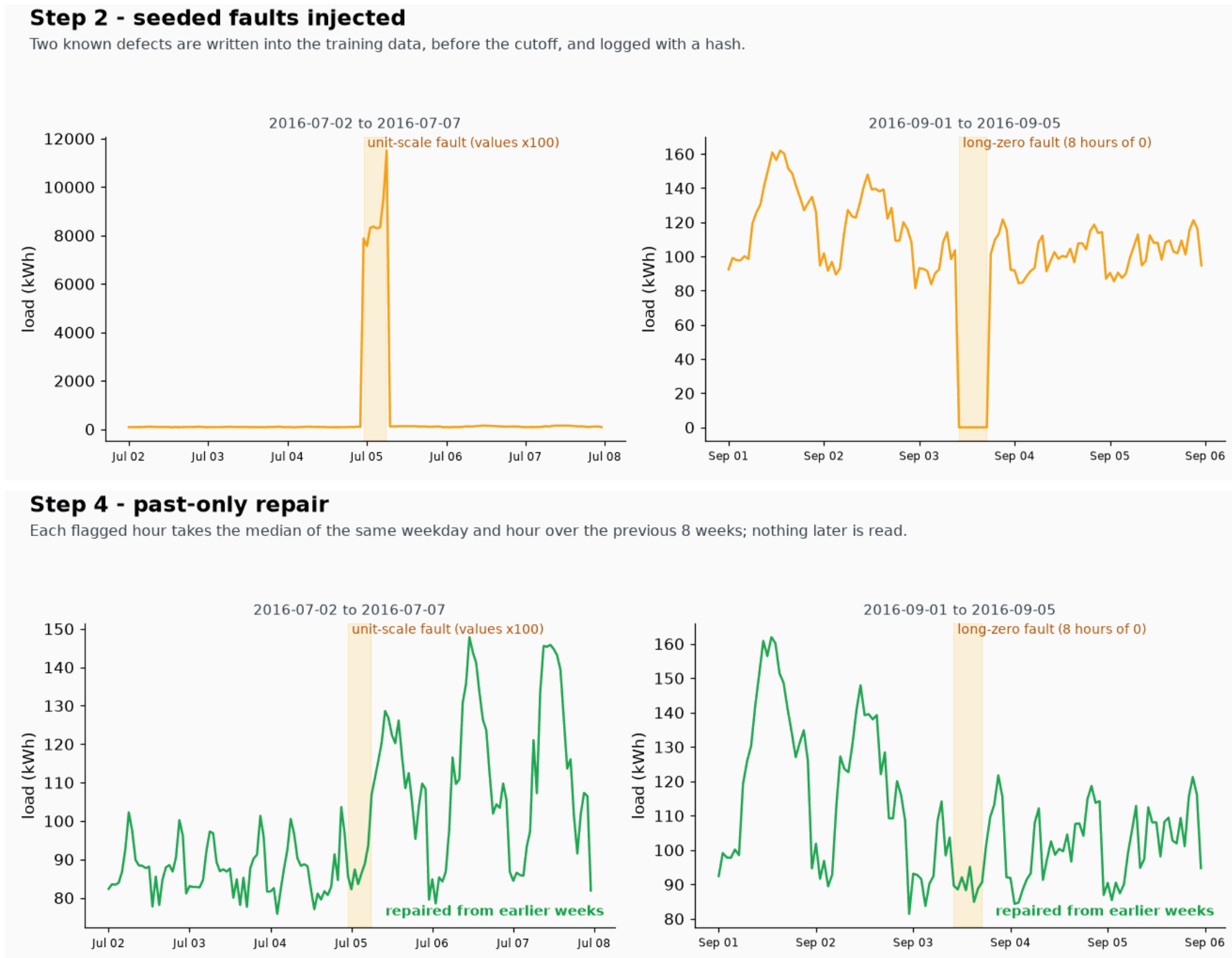

**Figure 4.** Real 2016 load of an education building at the Panther site. Top: the seeded ×100 unit error (July) and 8-hour zero block (September) in the corrupted condition. Bottom: the same windows after detection and past-only repair; each flagged hour took the median of the same weekday and hour over the previous eight weeks. Gradient-boosting MAE for this building: 11.1 kWh clean, 19.2 corrupted, 10.4 repaired.

### 5.5 The negative result that came first, and its diagnosis

The first pre-specified run (v2: fixed run-detector thresholds, forward fill only, one seeded event per family) reported the opposite of Table 1: the remediated condition forecast 25% *worse* than the corrupted one (MASE 0.922 versus 0.740). Because every artifact was published, the cause could be traced without re-running anything. Of the 22,033 rows the v2 remediation quarantined, 22,014 were natural data and only 19 were seeded faults. One coarsely quantized meter, whose readings repeat for hours as a matter of resolution, lost 97.9% of its training year to the stuck-sensor rule; buildings that are genuinely idle at night and at weekends lost those hours to the zero-run rule; and because only the tail of a run was flagged, the forward fill copied a run's own value back into it. The v3 amendment calibrates the run detectors per building on clean history, decides zero-run exemptions per run, flags whole runs, adds the hour-of-week profile repair and refuses to fill a flagged zero with a zero. Figure 5 shows the effect with the same model and split.

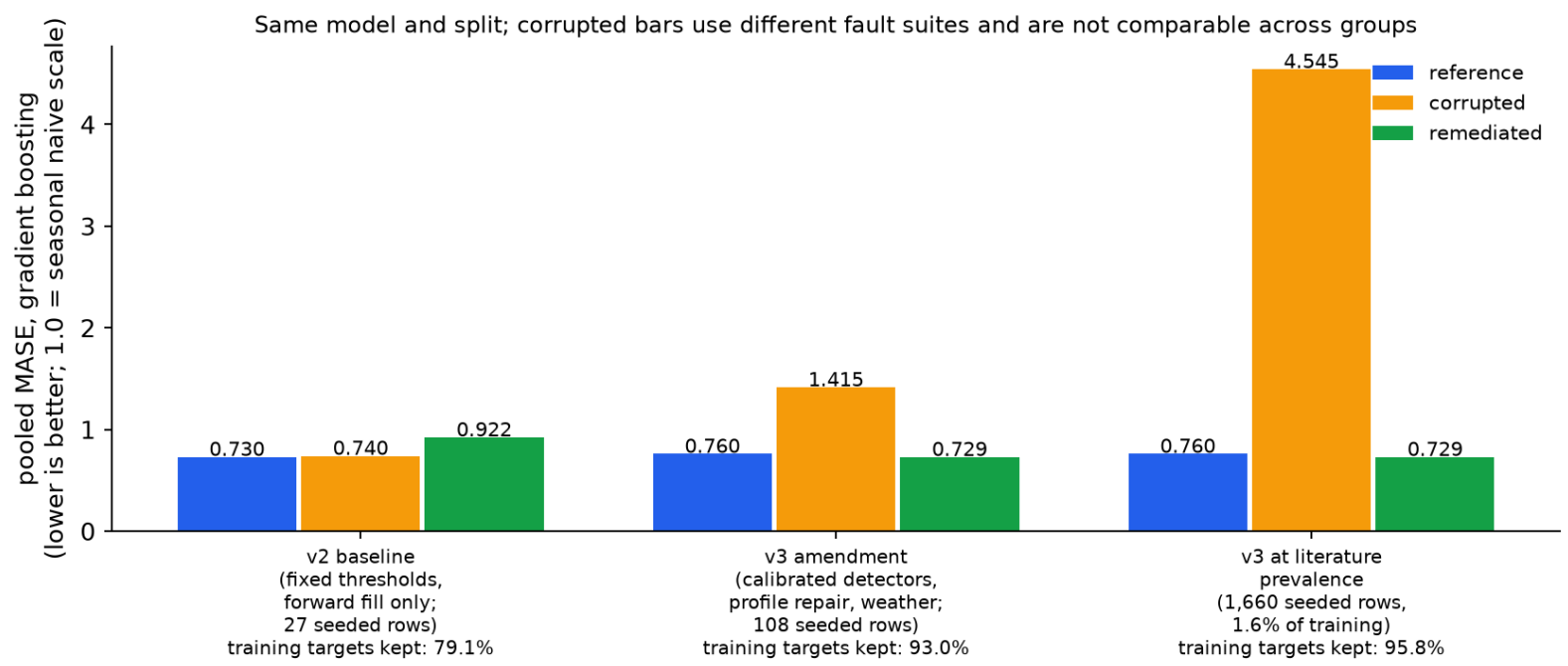

**Figure 5.** Remediation effect before and after the calibration, and at literature-level prevalence. The corrupted bars are not comparable across groups because the seeded suites differ (27, 108 and 1,660 rows); compare the clean and repaired bars and the share of training targets kept.

## 6. Limitations and threats to validity

- **Scope.** Twelve buildings at three sites in one time zone, two years, hourly resolution; a retrospective public-data benchmark, not an operational system. No claim of population representativeness is made.
- **Uncertainty.** The intervals in Tables 1 and 2 resample individual test hours and are likely too narrow because hourly errors are correlated; the block-bootstrap sensitivity run gives wider intervals. One pooled model serves all buildings, so a fault on one building moves every building's error.
- **Detector precision.** The checks flag many natural readings (for example 11,753 hours of natural zero runs in the sparse run). The source data carry no ground truth for natural anomalies, so those are counted as false positives by convention; they are the rows behind the 7.0% (sparse) and 4.2% (dense) of training targets that were quarantined.
- **Undetectable by construction.** Three meters have natural constant runs longer than eight hours, so their calibrated stuck thresholds (62, 73 and 1,371 hours) make an 8-hour seeded stuck segment undetectable; this costs one of four such events in the detector-validation suite and two of four in the forecasting condition (16 of the 108 seeded rows pass through unrepaired).
- **Weather.** Adding lagged weather and a fuller calendar (14 to 30 features, hyperparameters unchanged) made the clean-data gradient-boosting error slightly worse (0.730 to 0.760, +4%). Which features are responsible has not been isolated; next-day load likely depends on next-day weather, which this protocol deliberately withholds.
- **Sequence of versions.** v3 was designed two hours after the v2 result was examined; it is a dated, separately labeled second run, not a confirmation of a registered hypothesis. The v2 files are kept unchanged so the sequence can be audited.

## 7. Relevance and planned work

The parts of this work that transfer directly are the data contract, the per-meter calibrated checks, the hash-bound provenance chain and the three-condition evaluation design. Any organization that trains models on meter or sensor streams, whether a utility forecasting group, a building-energy analytics team or a grid operator's data platform, can reuse that design to measure whether its own cleaning rules help or hurt before those rules touch a production model, which is precisely the failure the v2 run exhibited and caught. The design serves the data-quality foundation that the Genesis Mission's grid challenge and the AI Action Plan's dataset agenda both presuppose [1, 2]. Planned next steps, in order: an adapter for hourly balancing-authority demand from the U.S. Energy Information Administration's Form EIA-930 so that the same checks run on transmission-scale data; more Building Data Genome 2 sites and a second time zone; forecast-weather inputs as a separately labeled experiment; and a streaming ingestion interface with the same contracts.

## 8. Reproducibility statement

All code, configurations, protocol files, amendments and eight result bundles are public at the repository above. Source data are downloaded from their pinned URLs and verified by size and SHA-256 before any run. Eighty-four automated tests and the hash verification of every committed bundle run in continuous integration on each push. With CPython 3.12.14 and the pinned package versions on the author's machine, a fresh run reproduced the onboarding outputs and the seasonal-naive, ridge and gradient-boosting predictions exactly; random-forest predictions match to about $10^{-12}$ because its trees are summed in thread order. Cross-platform byte identity is not claimed; CI recomputes the metrics of every committed bundle on Linux.

## References

[1] U.S. Department of Energy, Office of the Under Secretary for Science. *Genesis Mission National Science and Technology Challenges*, including "Scaling the Grid to Power the American Economy" (initial 26 challenges announced February 12, 2026; expanded to 33 in July 2026). https://www.energy.gov/undersecretaryforscience/genesis-mission/genesis-mission-national-science-and-technology-challenges (accessed September 28, 2026).

[2] The White House. *Winning the Race: America's AI Action Plan*, July 2025. Pillar I, "Build World-Class Scientific Datasets," pp. 8–9. https://www.whitehouse.gov/wp-content/uploads/2025/07/Americas-AI-Action-Plan.pdf.

[3] M. Weber, M. Turowski, H. K. Çakmak, R. Mikut, U. Kühnapfel, V. Hagenmeyer. "Data-driven copy-paste imputation for energy time series." *IEEE Transactions on Smart Grid* 12(6), 5409–5419, 2021. https://doi.org/10.1109/TSG.2021.3101831.

[4] J. Park, E. Alvarenga, J. Jeon, R. Li, F. Petropoulos, H. Kim, K. Ahn. "Probabilistic forecast-based portfolio optimization of electricity demand at low aggregation levels." *Applied Energy* 353, 122109, 2024. https://doi.org/10.1016/j.apenergy.2023.122109.

[5] E. Zapata-Webborn, J. Few, C. Hanmer. *Smart Meter Consumption Data: Data Quality Report*, edition 07. Smart Energy Research Lab, University College London, September 2024. Documentation for UK Data Service SN 8666.

[6] K. M. Fowler, A. H. Colotelo, J. L. Downs, K. D. Ham, J. W. Henderson, S. A. Montgomery, S. A. Parker, C. R. Vernon. *Simplified Processing Method for Meter Data Analysis*. Pacific Northwest National Laboratory, PNNL-24331, November 2015. https://doi.org/10.2172/1255411.

[7] E. Garrison and J. New. "Quality control methods for advanced metering infrastructure data." *Smart Cities* 4(1), 195–203, 2021. https://doi.org/10.3390/smartcities4010012.

[8] J. Shishido. "Smart meter data quality insights." *Proceedings of the 2012 ACEEE Summer Study on Energy Efficiency in Buildings*, pp. 12-277–12-288. American Council for an Energy-Efficient Economy, 2012.

[9] M. Gulati and P. Arjunan. "LEAD1.0: A large-scale annotated dataset for energy anomaly detection in commercial buildings." *Proceedings of the 13th ACM International Conference on Future Energy Systems (e-Energy '22)*, 485–488, 2022. https://doi.org/10.1145/3538637.3539761.

[10] C. Miller, A. Kathirgamanathan, B. Picchetti, P. Arjunan, J. Y. Park, Z. Nagy, P. Raftery, B. W. Hobson, Z. Shi, F. Meggers. "The Building Data Genome Project 2, energy meter data from the ASHRAE Great Energy Predictor III competition." *Scientific Data* 7, 368, 2020. https://doi.org/10.1038/s41597-020-00712-x. Data: https://doi.org/10.5281/zenodo.3887306.

[11] J. R. Schofield, R. Carmichael, S. H. Tindemans, M. Bilton, M. Woolf, G. Strbac. *Low Carbon London Project: Data from the Dynamic Time-of-Use Electricity Pricing Trial, 2013*. UK Data Service SN 7857, 2015. https://doi.org/10.5255/UKDA-SN-7857-1.

#### Appendix A. Artifacts behind each number

Tables 1 and 2, Figures 2 and 5: `metrics.csv`, `paired_differences.csv`, `building_mase.csv` in `energy-demand-forecasting/results/canonical_24h_v3/`, `canonical_24h_v3_dense/`, `canonical_24h_v2_final/`. Figure 3: `direct_1_to_24_sensitivity_v3/horizon_metrics.csv`. Detector metrics, thresholds and quarantine: `energy-ai-data-onboarding/results/bdg2_v3/`, `bdg2_v3_dense/`, `bdg2_mvp_v2_hardened/`.